\pdfoutput=1

\documentclass[11pt]{article}
\usepackage{colortbl}
\usepackage[table]{xcolor}
\usepackage[final]{acl}

\usepackage{times}
\usepackage{latexsym}

\usepackage[T1]{fontenc}

\usepackage[utf8]{inputenc}

\usepackage{microtype}

\usepackage{inconsolata}

\usepackage{graphicx}
\usepackage{times}
\usepackage{latexsym}
\usepackage[T1]{fontenc}
\usepackage[utf8]{inputenc}
\usepackage{pifont}
\newcommand{\cmark}{\ding{51}}%
\newcommand{\xmark}{\ding{55}}%

\usepackage{soul}
\usepackage{microtype}
\usepackage{tabularx}
\usepackage{xspace}
\usepackage{makecell}
\usepackage{tipa}
\usepackage{graphicx}
\usepackage{booktabs}
\usepackage{multirow}
\usepackage{caption}
\usepackage{amsmath} 
\usepackage{arydshln}
\usepackage{amssymb}  
\usepackage{xcolor}   
\usepackage{tcolorbox}
\usepackage{tabularx}
\usepackage{color}

\title{LDIR: Low-Dimensional Dense and Interpretable Text Embeddings \\with Relative Representations}

\author{
  Yile Wang, Zhanyu Shen, Hui Huang\thanks{Corresponding author.}   \\
  College of Computer Science and Software Engineering, Shenzhen University\\
  \texttt{wangyile@szu.edu.cn, zhanyushen.s@gmail.com, hhzhiyan@gmail.com}
}

\begin{document}
\maketitle
\begin{abstract}
Semantic text representation is a fundamental task in the field of natural language processing. Existing text embedding (e.g., SimCSE and LLM2Vec) have demonstrated excellent performance, but the values of each dimension are difficult to trace and interpret. Bag-of-words, as classic sparse interpretable embeddings, suffers from poor performance. Recently, \citet{benara2024crafting} propose interpretable text embeddings using large language models, which forms ``0/1'' embeddings based on responses to a series of questions. These interpretable text embeddings are typically high-dimensional (larger than 10,000). In this work, we propose \underline{L}ow-dimensional (lower than 500) \underline{D}ense and \underline{I}nterpretable text embeddings with \underline{R}elative representations (LDIR). The numerical values of its dimensions indicate semantic relatedness to different anchor texts through farthest point sampling, offering both semantic representation as well as a certain level of traceability and interpretability. We validate LDIR on multiple semantic textual similarity, retrieval, and clustering tasks. Extensive experimental results show that LDIR performs close to the black-box baseline models and outperforms the interpretable embeddings baselines with much fewer dimensions. Code is available at \url{https://github.com/szu-tera/LDIR}.

\end{abstract}

\section{Introduction}
Text embedding is an important technique for natural language processing by transforming textual data into numerical representations that capture semantics. The embeddings encode the meaning of contexts in a vector space where similar texts are close together in the representation space, playing vital roles in multiple tasks such as semantic textual similarity~\cite{agirre-etal-2012-semeval}, information retrieval~\cite{karpukhin-etal-2020-dense}, and retrieval-augmented generation~\cite{lewis2020retrieval}.

Seminal text embedding models mainly fall into two categories. The first is based on lightweight pre-trained models such as BERT~\cite{devlin-etal-2019-bert}, including Sentence-BERT~\cite{reimers-gurevych-2019-sentence}, SimCSE~\cite{gao-etal-2021-simcse}, etc. The second is based on current large language models such as LLaMA-3~\cite{dubey2024llama}, including LLM2Vec~\cite{behnamghader2024llmvec}, AngIE~\cite{li-li-2024-aoe}, etc. The text embeddings usually have 768 to 4096 dimensions correlated with the hidden size of the model and perform well across tasks. However, the values on each dimension are difficult to trace and interpret directly.

\begin{table}[t!]
	\centering
 \scalebox{0.62}{
	\begin{tabular}{lccc}
	\toprule
        \textbf{Embeddings}& \textbf{Type}& \textbf{Interp.}  & \textbf{Examples}\\
        \midrule
        SimCSE &Dense Emb& \multirow{2}*{\textcolor{red}{\xmark}} & 0.71, -0.03, ..., 0.13\\
        LLM2Vec& (768$\sim$4,096 dim.)& & (uninterpretable)\\
        \midrule
        \multirow{2}*{Bag-of-Words}&Sparse Emb& \multirow{2}*{\textcolor{teal}{\cmark}} & 12, 5, ..., 0\\
        &($\sim$30,000 dim)&&(occurrence of \texttt{<word>}$_i$)\\
        \midrule
        QAEmb-MBQA & 0/1 Emb&\multirow{2}*{\textcolor{teal}{\cmark}}  & 1, 0, ..., 1\\
        CQG-MBQA & ($\sim$10,000 dim.)& &(``yes/no'' to \texttt{<question>}$_i$)\\
        \midrule
        \multirow{2}*{\textbf{LDIR (Ours)}}&\textbf{Dense Emb} &\textbf{\multirow{2}*{\textcolor{teal}{\cmark}}}&\textbf{0.16, 0.83, ..., 0.35} \\
        &\textbf{($\sim$500 dim.)}&&(\textbf{relatedness to \texttt{<anchor>}$_i$})\\
        \bottomrule
\end{tabular}}
\caption{Comparing existing text embedding with LDIR. In particular, our text embedding is relatively low-dimensional and dense, while maintaining interpretable.}
\label{table:intro}
\end{table}

In contrast, bag-of-words is a classical sparse and interpretable text representation where each dimension's value represents the frequency of a specific word. However, its representational capability is limited. Recent QAEmb-MBQA~\cite{benara2024crafting} and CQG-MBQA~\cite{sun2024general} generate interpretable text embeddings based on large language models, where the value on each dimension is restricted to ``0/1'', reflecting ``yes/no'' answers to different crafted questions, as shown in Table~\ref{table:intro}. These works pave the way for new developments in interpretable text embedding.

In the aforementioned interpretable text embeddings, the value on each dimension are explicitly defined. However, due to the limitations of their sparse or ``0/1'' representations, they often require a high dimensionality. For example, bag-of-words is associated with the word vocabulary size, which typically reaches around 30K sizes in current models. QAEmb-MBQA and CQG-MBQA rely on designing a large number of questions, usually around 10K. In this work, we aim to mitigate the limitations of sparse or ``0/1'' representation and the requirement of high dimensionality.

Inspired by \citet{moschella2023relative}, we propose building low-dimensional, dense, and interpretable text embeddings with relative representations (LDIR). Specifically, the value on each dimension represents the relatedness to different ``anchor texts'' by automatic farthest point sampling, which can be floating-point numbers instead of ``0/1'', thereby leading a overall flexible and dense embeddings. We hope the dimensionality of vectors can be significantly reduced in this way, achieving a balance among semantic expressiveness, representational efficiency, and interpretability.

We conduct extensive experiments to validate the semantic expressiveness of LDIR, including evaluations on seven semantic textual similarity tasks, six retrieval tasks, and seven clustering tasks. Experimental results show that our embeddings, with only 500 embedding dimensions, achieve comparable performance to black-box embeddings, and outperforms interpretable embedding baselines of QAEmb-MBQA and CQG-MBQA.

\section{Related Work}

\textbf{Semantic Text Embedding.} For pre-trained neural model based text embeddings, Sentence-BERT~\cite{reimers-gurevych-2019-sentence} and SimCSE~\cite{gao-etal-2021-simcse} use siamese network or contrastive learning to improve the semantic representations of BERT~\cite{devlin-etal-2019-bert}. For large language model based text embeddings, MetaEOL~\cite{lei-etal-2024-meta} elicit the text embeddings from a single-token through prompting LLaMA~\cite{touvron2023llama}. AngIE~\cite{li-li-2024-aoe} optimizes angle differences instead of cosine similarity for better modeling semantics. LLM2Vec~\cite{behnamghader2024llmvec} enables bidirectional attention and avoid the limitations of decoder-only models in encoding texts. NV-Embed~\cite{lee2025nvembed} propose latent attention layer to enhance the general-purpose embedding tasks of decoder-only models. Although these methods produce dense text embedding and show strong performance on benchmarks, however, it's difficult to interpret each dimension of the resulted dense representations. 

\noindent\textbf{Interpretable Text Embedding.} Bag-of-words (BoW) is one of the most classic and interpretable text embedding, where each dimension represents the frequency of occurrence for each word in the vocabulary. However, its representation capacity is largely limited. LISA~\cite{patel-etal-2023-learning} distills linguistically interpretable style attribute embeddings for author style representation, which is not general semantics oriented. QAEmb-MBQA~\cite{benara2024crafting} proposes designing multiple questions and use the ``yes/no'' answers by language models as ``0/1'' embeddings for semantic representations. CQG-MBQA~\cite{sun2024general} further improves the question generation module and reduce the costs for obtaining the ``0/1'' embeddings. Unlike these approaches, we develop low-dimensional dense text embeddings which can improve the general semantic expressiveness while preserving a certain of interpretability.

\noindent\textbf{Relative Representation.} \citet{moschella2023relative} observe that the latent similarity between each sample and a fixed set of anchors in neural networks shows invariance for different training settings. They call such latent similarity as relative representation and verify the findings on the space of word embeddings~\cite{mikolov2013efficient,bojanowski-etal-2017-enriching} or model stitching~\cite{bansal2021revisiting}. \citet{norelli2023asif} use relative representations to align images and text for zero-shot classification.  \citet{Maniparambil_2024_CVPR} also leverage such cross-modal alignment for caption matching and retrieval tasks. Overall, the above studies primarily illustrate the potential of relative representation for feature alignment, rather than utilizing it for sentence-level semantic representations. 

\section{Method}

\begin{figure*}[t]
    \centering
    \includegraphics[scale=1.05]{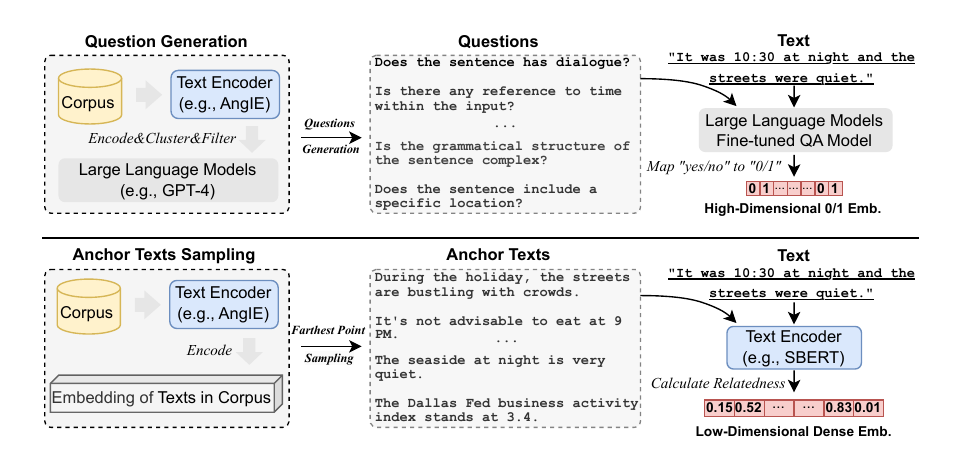}
    \caption{Comparison between baselines (top) and our method (bottom). \citet{benara2024crafting} and \citet{sun2024general} uses the generated questions and ``yes/no'' answers to build high-dimensional ``0/1''  text embeddings. We use anchor texts via farthest point sampling to calculate the relatedness and build low-dimensional dense text embeddings.}
    \label{figure:method}
\end{figure*}

\subsection{QA-Embedding Baseline}
Formally, we define the semantic text embedding $e(t)$ of a given a text $t$ as:
\begin{equation}
e(t) = \mathcal{E}(t) = [v_1, v_2, ..., v_d],
\end{equation}
where $\mathcal{E}$ is an embedding model, $v_1, v_2, ..., v_d$ are the values in each dimension. For neural model $\mathcal{E}$, the embedding $[v_1, v_2, ..., v_d]$ generally encode some high-level semantic features. However, it is difficult to interpret the specific meanings of each value $v_i$ directly.

\citet{benara2024crafting} propose generating interpretable QA-Embedding to make the values in each dimension meaningful. In particular, they first generate $k$ questions $q_1, q_2, ..., q_k$ by leveraging GPT-4 with specific prompts, then obtain the $k$ ``yes'' or ``no'' responses $\{r_j\}_{j=1}^k$ by using large language models $\mathcal{M}$ (e.g., GPT-4 or LLaMA-3):
\begin{equation}
r_1, ..., r_k = \mathcal{M}(q_1\oplus t), ..., \mathcal{M}(q_k\oplus t),
\end{equation}
where $\oplus$ indicates concatenation operation, and the answer $r_j$ is either ``yes'' or ``no'' to the question $q_j$. 

The interpretable text embedding is defined as the ``0/1'' sequence of according to the answers:
\begin{equation}
\begin{array}{l}
\begin{aligned}
&e_\text{0/1}^\text{interp}(t) =  [\hat{r}_1, \hat{r}_2, ..., \hat{r}_k],\\
&\hat{r}_j =
\begin{cases} 
1,  & \mbox{if}\ r_j=\mathcal{M}(q_j\oplus t)\ \text{is}\ \text{``yes''},\\
0, & \mbox{otherwise}.
\end{cases}
\end{aligned}
\end{array}
\label{qaemb}
\end{equation}

\citet{sun2024general} further improves the QA-Embedding by reducing the costs of calling LLMs through applying external trainable question generation module. We show an example of QA-Embedding in the top of Figure~\ref{figure:method}.

Both \citet{benara2024crafting} and \citet{sun2024general} rely on high-quality questions $\{q_j\}_{j=1}^k$ generated via GPT-4 or trained models, followed by extra filtering. On the other hand, since $\hat{r}_j$ can only be 0 or 1, the number of questions is extremely large, often reaching 9000$\sim$10000 (i.e., the number $k$).

\subsection{``0/1'' Embedding to Dense Embedding}
Although the above values $\{\hat{r}_j\}_{j=1}^k$ of the embedding $e^\text{interp}(t)$ is interpretable, the expressive power is greatly limited because all values are either 0 or 1. Therefore, we consider using dense representations with floating-point number of each dimension. The overview of our dense and interpretable embedding is illustrated in the bottom of Figure~\ref{figure:method}.

Instead of generating $k$ questions $q_1, q_2, ..., q_k$, we first select $n$ representative texts $a_1, a_2, ..., a_n$ which we called ``anchor texts'' (detailed in the following subsection). Then, for each anchor text $a_j$, we compute the relatedness between $a_j$ and $t$.
\begin{equation}
s_j = \textsc{Rel}(a_j, t),
\end{equation}
where $\textsc{Rel}(a_j, t)$ indicates the relatedness score of $a_j$ and $t$, which can be computed through general encoder or trained text encoders using widely used cosine similarity~\cite{mikolov2013efficient,reimers-gurevych-2019-sentence, ethayarajh-2019-contextual}:
\begin{equation}
\begin{array}{l}
\begin{aligned}
&\textsc{Rel}(a_j, t)  = \frac{\textsc{Enc}(a_j)\cdot \textsc{Enc}(t)}{||\textsc{Enc}(a_j)||\cdot ||\textsc{Enc}(t)|| },\\
\end{aligned}
\end{array}
\label{relcal}
\end{equation}
where $\textsc{Enc}$ can be an pre-trained encoder such as SimCSE~\cite{gao-etal-2021-simcse}, ModernBERT~\cite{modernbert}, and AngIE~\cite{li-li-2024-aoe}, without further fine-tuning.

After obtaining the relatedness scores $\{s_j\}_{j=1}^n$,  our dense and interpretable embeddings is:
\begin{equation}
\begin{array}{l}
\begin{aligned}
e_\text{dense}^\text{interp}(t) &=[s_1, ..., s_n]\\
&=  [\textsc{Rel}(a_1, t), ..., \textsc{Rel}(a_n, t)].\\
\end{aligned}
\end{array}
\label{relemb}
\end{equation}

Comparing Eq.~\ref{relemb} with Eq.~\ref{qaemb}, our embeddings $e_\text{dense}^\text{interp}(t)$ are represented based on floating-point numbers, which can alleviate the requirement for high dimensionality. In practice, we find that 200$\sim$500 (i.e., the number $n$) anchor texts are enough to achieve better performance.

\subsection{Anchor Texts Selection}
\label{anchor}

In contrast to other interpretable embedding methods that rely on the extensive general knowledge of large language models (e.g., GPT-4) to derive and filter explanatory questions, we propose to automatically sample anchor texts from the corpus without generation and filtering. Selecting representative texts $a_1, a_2, ..., a_n$ as anchor texts are important. For example, if all the anchor texts are similar to each other, then values in the final embeddings in Eq.~\ref{relemb} will also be close, thus the representativeness of overall embedding is limited. In practice, we use the farthest point sampling algorithm to select the representative anchor texts and compare with other sampling methods such as uniform sampling and K-Means in Section \ref{compare_anchor_selection}.

\begin{figure}[t]
    \centering
    \includegraphics[scale=0.7]{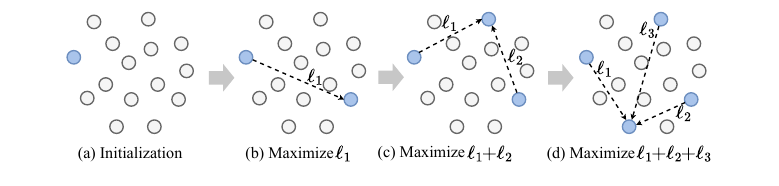}
    \caption{Example of extracting four anchor texts using farthest point sampling. Each dot represents a text embedding through an encoder, with the blue dots indicating the extracted anchor texts.}
    \label{figure:fps}
\end{figure}

Farthest point sampling (FPS) is a widely used method that identifies representative subset within the complete feature space. In our workflow, the generation process begins by applying an encoder to the corpus texts, producing $N$ embeddings $\{e_1, e_2, ..., e_N\}$. Then a subset $\{\hat{e}_1, \hat{e}_2, ..., \hat{e}_n\}$ is sampled gradually such that the distance between any two embedding vectors $(\hat{e}_i, \hat{e}_j)$ are farther apart than that of any embedding vector pair in the remaining set $\{e_1, e_2, ..., e_{N-n}\}$. Therefore, the text corresponding to the collected subset of embeddings constitutes the anchor texts $a_1, a_2, ..., a_n$ required. An example of FPS procedure is shown in Figure~\ref{figure:fps}. \citet{benara2024crafting} uses AngIE~\cite{li-li-2024-aoe} to encode texts in corpus for generating questions by GPT-4. We follow them and set AngIE as the text encoder.

\subsection{Overall Comparison}
We make an overall comparison to existing embedding methods in Table~\ref{table:overall_comparison}. Comparing from different perspectives, we achieve \textbf{low-dimensional and dense embeddings with low costs}. Specifically, SimCSE and LLM2Vec require additional contrastive learning to fine-tune models, QAEmb-MBQA and CQG-MBQA require extra calls on GPT-4 or training question generation modules. As comparison, LDIR primarily employs a automatically farthest point sampling strategy to extract anchor texts, and the calculation of embeddings does not require prompting large language models.

We also notice a \textbf{trade-off between interpretability and performance} across embeddings. For example, SimCSE and LLM2Vec exhibit strong semantic representation capabilities, but their interpretability is relatively low. On the other hand, bag-of-words offers high interpretability but performs poorly. Compared to QAEmb-MBQA and CQG-MBQA, LDIR sacrifices some interpretability (as the overall semantic relatedness is less direct and explicit than responses to questions) but achieves better performance, comparable to embeddings like SimCSE (see experimental results in Section~\ref{sec:results}). In Section~\ref{sec:relatedness}, we further discuss how to obtain fine-grained relatedness to enhance the interpretability of our approach.

\begin{table}[t!]
	\centering
 \scalebox{0.61}{
	\begin{tabular}{lccccc}
	\toprule
        \textbf{Embeddings}& \textbf{Type}& \textbf{Dim.}$\downarrow$& $^\dag$\textbf{Ext. Costs}$\downarrow$ & \textbf{Interp.}$\uparrow$ & \textbf{Perf.}$\uparrow$\\
        \midrule
        SimCSE &\multirow{2}*{Dense}&\multirow{2}*{\textcolor{brown}{Medium}}& \multirow{2}*{\textcolor{brown}{Medium}}& \multirow{2}*{{\textcolor{red}{Low}}}&\multirow{2}*{\textcolor{teal}{High}}\\
        LLM2Vec&& && &\\
        \midrule
        \multirow{2}*{Bag-of-Words}&\multirow{2}*{Sparse}& \multirow{2}*{{\textcolor{red}{Large}}}&\multirow{2}*{{\textcolor{teal}{Low}}}& \multirow{2}*{{\textcolor{teal}{High}}}&\multirow{2}*{{\textcolor{red}{Low}}}\\
        &&&&&\\
        \midrule
        QAEmb-MBQA & \multirow{2}*{0/1 Emb.}&\multirow{2}*{{\textcolor{red}{Large}}} &\multirow{2}*{{\textcolor{red}{High}}}&  \multirow{2}*{{\textcolor{teal}{High}}}&\multirow{2}*{\textcolor{brown}{Medium}}\\
        CQG-MBQA & && &&\\
        \midrule
        \multirow{2}*{{LDIR (Ours)}}&\multirow{2}*{{Dense}}&\multirow{2}*{{\textcolor{teal}{Small}}}&\multirow{2}*{{{\textcolor{teal}{Low}}}}& \multirow{2}*{{\textcolor{brown}{Medium}}}&\multirow{2}*{{\textcolor{teal}{High}}}\\
        &&&&&\\
        \bottomrule
\end{tabular}}
\caption{Overall comparison between existing embeddings and LDIR. $\dag$: The external costs indicate the costs for external training or calling large language models.}
\label{table:overall_comparison}
\end{table}

\begin{table*}[t!]
    \centering
    \scalebox{0.69}{
    \begin{tabular}{lrccccccccc}
        \toprule
        \multirow{2.5}*{\textbf{Model}} & \multirow{2.5}*{\textbf{Dim.}}  &\multirow{2.5}*{\textbf{Type}}&\multicolumn{8}{c}{\textbf{Spearman Correlation (STS)}}\\
        \cmidrule(lr){4-11}
        &&& \textbf{STS12} & \textbf{STS13} & \textbf{STS14} & \textbf{STS15} & \textbf{STS16} & \textbf{STS-B} & \textbf{SICK-R} & \textbf{Avg.}\\
        \midrule
        \multicolumn{11}{c}{(\textit{Black-Box Embeddings})}\\
        BERT$_\text{base}$~\cite{devlin-etal-2019-bert} & 768&Dense & 38.78 & 57.98 & 57.98 &63.15 & 61.06 & 46.35&58.40&54.81 \\
        GloVe~\cite{pennington-etal-2014-glove} &300 &Dense & 54.64 & 69.16 & 60.81 & 72.31& 65.34 & 61.54 &55.43&62.74\\
        USE~\cite{cer-etal-2018-universal} &512 &Dense & 64.49 & 67.80 & 64.61 &76.83 & 73.18 & 74.92 &76.69&71.22\\
        LLM2Vec~\cite{behnamghader2024llmvec} &4,096 &Dense &61.60 &79.71 &72.11 &82.18 &79.41 &77.44 &72.16 &74.94 \\
        SimCSE$_\text{unsup}$~\cite{gao-etal-2021-simcse} & 768&Dense & 66.05 & 81.49 &73.61  & 79.72&78.12  & 76.52 &72.24&75.39\\
        SBERT$_\text{ori}$~\cite{reimers-gurevych-2019-sentence} & 768&Dense & 74.53 & 77.00 & 73.18 & 81.85& 76.82 & 79.10 &74.29&76.68\\
        SimCSE$_\text{sup}$~\cite{gao-etal-2021-simcse} &768 &Dense &75.30  &84.67  & 80.19 &85.40 & 80.82 &84.25  &68.38&79.86\\
        WhitenedCSE~\cite{zhuo-etal-2023-whitenedcse} &768 &Dense &74.65  & 85.79 &77.49  &84.71 &80.33  &  81.48&75.34&79.97\\
        SBERT$_\text{new}$~\cite{reimers-gurevych-2019-sentence} & 768&Dense & 73.08 & 82.13 & 76.73 & 85.58& 80.23 & 83.09 &79.32&80.02\\
        
        text-embedding-ada-002~\cite{openaiapi} &1,536 &Dense &72.84 &86.10  & 81.15 & 88.49& 85.08 & 83.56 &79.00&82.31\\
        ModernBERT$_\text{emb}$~\cite{modernbert} & 1,024&Dense &80.67 &87.87 &83.80 &88.59 &86.82 &87.40 &80.31 &85.07\\
        AngIE~\cite{li-li-2024-aoe} &1,024 &Dense & 79.09 &89.62  & 85.02 & 89.51& 86.61 & 89.06 &82.62&85.93\\
        \midrule
        \multicolumn{11}{c}{(\textit{Interpretable Embeddings})}\\
        Bag-of-Words (with BERT vocabulary)&28,996 &Sparse & 44.75 & 52.06 & 54.78 &68.65 & 60.59 & 54.85 &57.87&56.22\\
        QAEmb-MBQA~\cite{benara2024crafting} &10,654 &0/1 Emb. &59.40  & 63.19 & 57.68 &69.29 & 63.18 & 71.33 &72.33&65.20\\
        CQG-MBQA~\cite{sun2024general} &9,614 &0/1 Emb. & 69.21 &80.19  & 73.91 & 80.66& 78.30 & 82.69 &78.21&77.60\\
        
        \rowcolor{gray!20}
        LDIR (with SBERT$_\text{new}$)  &200 &Dense &74.44 &77.73 &73.58 &81.93 &77.98 &80.82 &78.42 &77.84 \\
        \rowcolor{gray!20}
        LDIR (with SBERT$_\text{new}$)  &500 &Dense &72.51 &79.92 &74.93 &82.93 &78.44 &81.88 &79.34 &78.55 \\
        \rowcolor{gray!20}
        LDIR (with ModernBERT$_\text{emb}$)  &200 &Dense &72.99 &79.63 &75.09 &82.52 &79.93 &82.61 &78.43 &78.74 \\
        \rowcolor{gray!20}
        LDIR (with ModernBERT$_\text{emb}$)  &500 &Dense &72.30 &81.97 &75.48 &82.86 &79.88 &83.22 &\underline{80.72} &79.49 \\
        \rowcolor{gray!20}
        LDIR (with AngIE)  &200 &Dense &\textbf{79.28} &\underline{82.16} &\textbf{80.96} &\underline{84.64} &\textbf{83.68} &\underline{85.32} &77.92 &\underline{81.99} \\
        \rowcolor{gray!20}
        LDIR (with AngIE)  &500 &Dense &\underline{78.85} &\textbf{84.35} &\underline{80.93} &\textbf{84.79} &\underline{83.61} &\textbf{86.31} &\textbf{80.85} &\textbf{82.82} \\
        
        \bottomrule
    \end{tabular}}
\caption{Comparison between baselines and LDIR on semantic textual similarity (STS) tasks. Among the interpretable embeddings, the best results in each column are \textbf{in bold}, and the second-best results are \underline{underlined}.}
\label{main_results_sts}
\end{table*}

\section{Experiments}

\subsection{Settings}

\noindent\textbf{Corpus.} We use the same texts resource MEDI2\footnote{\url{https://huggingface.co/datasets/GritLM/MEDI2}} and pre-processing steps (merging and filtering) by~\citet{sun2024general}, which results in a final corpus of approximately 6.8 million sentences for task-agnostic anchor texts extraction.

\noindent\textbf{Tasks and Datasets.} For semantic textual similarity (STS) tasks, we evaluate on SemEval STS tasks 2012-2016~\cite{agirre-etal-2012-semeval,agirre-etal-2013-sem,agirre-etal-2014-semeval,agirre-etal-2015-semeval,agirre-etal-2016-semeval}, STS Benchmark~\cite{cer-etal-2017-semeval}, and SICK-Relatedness~\cite{marelli-etal-2014-sick}.

For retrieval tasks, we use 1\% of the samples from the MS MARCO~\cite{bajaj2016ms} development set, ArguAna~\cite{wachsmuth-etal-2018-retrieval}, FiQA-2018 (FQA; ~\citealp{ACM-FiQA}), NFCorpus (NFC;~\citealp{springer-NFCorpus}), SCIDocs~\cite{cohan-etal-2020-specter}, and SciFact~\cite{wadden-etal-2020-fact}. 

For clustering tasks, we use TwentyNewsgroups (TNG), StackExchange (SE-P2P), Biorxiv (BR-P2P, BR-S2S), Medrxiv (MR-P2P, MR-S2S), and Reddit (RD-P2P) from massive text embedding benchmark (MTEB;~\citealp{muennighoff-etal-2023-mteb}).

\noindent\textbf{Baselines.} We use the baselines in \citet{sun2024general}, including {GloVe}~\cite{pennington-etal-2014-glove}, {USE}~\cite{cer-etal-2018-universal}, {BERT}~\cite{devlin-etal-2019-bert}, {SBERT}~\cite{reimers-gurevych-2019-sentence}, {SimCSE}~\cite{gao-etal-2021-simcse}, API-based {text-embedding-ada-002}~\cite{openaiapi}, {WhitenedCSE}~\cite{zhuo-etal-2023-whitenedcse}, {AngIE}~\cite{li-li-2024-aoe}, {Bag-of-Words}, {BM25}~\cite{robertson2009probabilistic}, {QAEmb-MBQA}~\cite{benara2024crafting}. We also use LLM2Vec~\cite{behnamghader2024llmvec}, ModernBERT~\cite{modernbert}, and {CQG-MBQA}~\cite{sun2024general} as additional baselines.

\noindent\textbf{Evaluations.} We use the official evaluation suite of BEIR~\cite{thakur2021beir} for MS MARCO and MTEB~\cite{muennighoff-etal-2023-mteb} for other datasets. Spearman correlation, nDCG@10, and V-measure metrics are used as metrics for STS, retrieval, and clustering tasks, respectively.

\begin{table*}[t!]
    \centering
    \scalebox{0.69}{
    \begin{tabular}{lrcccccccc}
        \toprule
        \multirow{2.5}*{\textbf{Model}} & \multirow{2.5}*{\textbf{Dim.}}  &\multirow{2.5}*{\textbf{Type}}&\multicolumn{7}{c}{\textbf{nDCG@10 (Retrieval)}}\\
        \cmidrule(lr){4-10}
        &&& \textbf{MS MARCO} & \textbf{ArguAna} & \textbf{FQA} & \textbf{NFC} & \textbf{SCIDocs} & \textbf{SciFact} & \textbf{Avg.}\\
        \midrule
        \multicolumn{10}{c}{(\textit{Black-Box Embeddings})}\\
        BERT$_\text{base}$~\cite{devlin-etal-2019-bert} & 768&Dense &16.86 &28.29  &\phantom{0}2.19 &\phantom{0}4.30  &\phantom{0}2.82 &13.34 &11.30\\
        SimCSE$_\text{unsup}$~\cite{gao-etal-2021-simcse} &768 &Dense &44.63 &38.34  &\phantom{0}9.84 &\phantom{0}9.88  &\phantom{0}5.50 &25.72 &22.32\\
        GloVe~\cite{pennington-etal-2014-glove} & 300&Dense &44.27 &36.30  &10.09 &13.87  &\phantom{0}8.04 &29.58 &23.69\\
        SimCSE$_\text{sup}$~\cite{gao-etal-2021-simcse} & 768&Dense &47.86 &39.33  &10.41 &12.42  &\phantom{0}7.53 &29.59 &24.52\\
        LLM2Vec~\cite{behnamghader2024llmvec} &4,096 &Dense &63.48 &51.73 &28.56 &26.29 &10.39 &66.36 &41.14 \\
        SBERT$_\text{new}$~\cite{reimers-gurevych-2019-sentence} & 768&Dense &88.74 &47.13  &37.27 &32.25  &21.82 &62.64 &48.31\\
        ModernBERT$_\text{emb}$~\cite{modernbert} &1,024 &Dense &87.39 &46.56 &45.18 &33.99 &21.40 &69.98 &50.75\\
        AngIE~\cite{li-li-2024-aoe} &1,024 &Dense &90.43  &66.15  &44.84 &38.65  &22.98 &74.07 &56.19\\
        text-embedding-ada-002~\cite{openaiapi} & 1,536&Dense &92.18  &58.05  &55.00 &42.07  &23.11 &77.77 &58.03\\
        
        \midrule
        \multicolumn{10}{c}{(\textit{Interpretable Embeddings})}\\
        Bag-of-Words (with BERT vocabulary)&28,996 &Sparse &29.79 &34.25  &\phantom{0}3.99 &21.51  &\phantom{0}6.79 &47.36 &23.95\\
        BM25~\cite{robertson2009probabilistic} &$^\dag$N/A &Sparse &68.42  &49.28  &25.14 &\textbf{32.08}  &15.78 &\textbf{68.70} &43.23\\
        QAEmb-MBQA~\cite{benara2024crafting} &10,654 &0/1 Emb. &40.51  &34.75  &\phantom{0}8.23 &\phantom{0}3.87  &\phantom{0}3.74 &12.01 &17.19\\
        CQG-MBQA~\cite{sun2024general} &9,614 &0/1 Emb. &62.21 &47.75  &18.63 &\phantom{0}9.74  &\phantom{0}8.67 &32.80 &29.97\\
        
        \rowcolor{gray!20}
        LDIR (with ModernBERT$_\text{emb}$)  &200 &Dense &57.65 &33.62 &21.06 &16.05 &11.27 &49.59 &31.54 \\
        \rowcolor{gray!20}
        LDIR (with ModernBERT$_\text{emb}$)  &500 &Dense &64.75 &37.91 &26.34 &18.06 &13.87 &52.13 &35.84 \\
        \rowcolor{gray!20}
        LDIR (with SBERT$_\text{new}$)  &200 &Dense &78.24 &42.25 &25.94 &25.35 &\underline{18.34} &55.42 &40.92 \\
        \rowcolor{gray!20}
        LDIR (with SBERT$_\text{new}$)  &500 &Dense &\underline{82.01} &44.89 &30.52 &\underline{26.06} &\textbf{19.74} &56.25 &43.24 \\
        \rowcolor{gray!20}
        LDIR (with AngIE) &200 &Dense &\textbf{83.11} &\underline{57.11} &\underline{30.86} &27.06 &15.44 &59.85 &\underline{45.57} \\
        \rowcolor{gray!20}
        LDIR (with AngIE) &500 &Dense &82.00 &\textbf{60.10} &\textbf{33.34} &24.67 &17.72 &\underline{62.26} &\textbf{46.68} \\
        \bottomrule
    \end{tabular}}
\caption{Comparison between baselines and LDIR on retrieval tasks. $\dag$: BM25 is a lexical retrieval algorithm and does not generate fixed-dimensional vectors.}
\label{main_results_retrieval}
\end{table*}

\begin{table*}[t!]
    \centering
    \scalebox{0.65}{
    \begin{tabular}{lrccccccccc}
        \toprule
        \multirow{2.5}*{\textbf{Model}} & \multirow{2.5}*{\textbf{Dim.}}  &\multirow{2.5}*{\textbf{Type}}&\multicolumn{8}{c}{\textbf{V-Measure (Clustering)}}\\
        \cmidrule(lr){4-11}
        &&& \textbf{TNG} & \textbf{SE-P2P} & \textbf{BR-P2P} & \textbf{BR-S2S} & \textbf{MR-P2P} & \textbf{MR-S2S} & \textbf{RD-P2P} & \textbf{Avg.}\\
        \midrule
        \multicolumn{11}{c}{(\textit{Black-Box Embeddings})}\\
        
        SimCSE$_\text{unsup}$~\cite{gao-etal-2021-simcse} &768 &Dense &23.21 &28.50  &24.90  &19.55 &23.60  &21.97 &45.14 &26.70\\
        GloVe~\cite{pennington-etal-2014-glove} & 300&Dense &25.83 &28.51  &29.27  &19.18 &26.12  &20.38 &35.82 &26.44\\
        BERT$_\text{base}$~\cite{devlin-etal-2019-bert} & 768&Dense &23.35 &26.55  &30.12  &24.77 &26.09  &23.60 &43.32 &28.26\\
        SimCSE$_\text{sup}$~\cite{gao-etal-2021-simcse} & 768&Dense &34.86 &29.45  &30.15  &24.67 &26.25  &24.12 &47.74 &31.03\\
        LLM2Vec~\cite{behnamghader2024llmvec} &4,096 &Dense &32.02 &36.36 &38.39 &31.31 &31.47 &27.87 &61.67 &37.01 \\
        SBERT$_\text{new}$~\cite{reimers-gurevych-2019-sentence} & 768 &Dense &47.47 &33.13  &36.99  &33.21 &34.25  &32.24 &54.80 &38.87\\
        ModernBERT$_\text{emb}$~\cite{modernbert} &1,024 &Dense &51.26 &34.73 &39.47 &34.67 &34.40 &31.63 &64.68 &41.55\\
        AngIE~\cite{li-li-2024-aoe} &1,024 &Dense &51.72 &36.72  &39.38  &37.23 &33.22  &31.18 &65.35 &42.11\\
        text-embedding-ada-002~\cite{openaiapi} &1,536 &Dense &58.14 &36.88  &38.03  &36.53 &32.70  &31.27 &67.96 &43.07\\
        
        \midrule
        \multicolumn{11}{c}{(\textit{Interpretable Embeddings})}\\
        Bag-of-Words (with BERT vocabulary)&28,996 &Sparse &\phantom{0}8.52 &17.64  &\phantom{0}4.70  &\phantom{0}3.32 &11.39  &13.05 &15.67 &10.61\\
        QAEmb-MBQA~\cite{benara2024crafting} &10,654 &0/1 Emb. &36.72 &25.68  &24.66  &21.16 &25.53  &22.85 &46.57 &29.02\\
        CQG-MBQA~\cite{sun2024general} &9,614 &0/1 Emb. &\underline{40.00} &28.22  &34.88  &31.13 &31.02  &28.71 &54.40 &35.48\\

        \rowcolor{gray!20}
        LDIR (with ModernBERT$_\text{emb}$)  &200 &Dense &29.72 &37.03 &25.93 &21.06 &31.12 &27.65 &52.65 &32.17 \\
        \rowcolor{gray!20}
        LDIR (with ModernBERT$_\text{emb}$)  &500 &Dense &30.18 &37.36 &28.40 &23.53 &32.42 &29.27 &52.77 &33.42 \\
        \rowcolor{gray!20}
        LDIR (with AngIE) &200 &Dense &36.93 &\textbf{40.60} &34.27 &31.61 &33.62 &32.84 &\textbf{58.88} &38.39 \\
        \rowcolor{gray!20}
        LDIR (with AngIE) &500 &Dense &37.63 &38.82 &34.82 &\underline{33.05} &34.79 &33.63 &\underline{57.92} &\underline{38.67} \\
        \rowcolor{gray!20}
        LDIR (with SBERT$_\text{new}$)  &200 &Dense &36.72 &\underline{39.44} &\underline{35.81} &30.81 &\underline{36.20} &\underline{33.98} &52.60 &37.94 \\
        \rowcolor{gray!20}
        LDIR (with SBERT$_\text{new}$)  &500 &Dense &\textbf{40.43} &37.86 &\textbf{38.07} &\textbf{33.32} &\textbf{37.63} &\textbf{35.63} &52.54 &\textbf{39.36} \\
        
        \bottomrule
    \end{tabular}}
\caption{Comparison between baselines and LDIR on clustering tasks.}
\label{main_results_clustering}
\end{table*}

\subsection{Main Results}
\label{sec:results}
We apply different encoders \textsc{Enc} in Eq.~\ref{relcal} and sample different number of anchor texts ($n=200$ and $n=500$) for comparison.

\noindent\textbf{Semantic Textual Similarity.} The results are shown in Table~\ref{main_results_sts}. LDIR demonstrates significant performance improvements among interpretable text embedding models while achieving competitive results compared to state-of-the-art black-box models. For example, LDIR outperforms the best interpretable embeddings CQG-MBQA (82.82 vs. 77.60), and also surpasses black-box models including SBERT$_\text{new}$ (80.02) and OpenAI’s text-embedding-ada-002 (82.31).

Comparing different dimensional settings, the 500-dimensional embeddings perform slightly better than the 200-dimensional ones. Among different backbone encoder models, AngIE gives the best results, while SBERT performs the worst, reflecting their differences in semantic encoding capabilities.

\noindent\textbf{Retrieval.} The results are shown in Table~\ref{main_results_retrieval}. LDIR achieves a nDCG@10 score of 46.68 with AngIE encoder and 500 dimensions, outperforming the traditional retrieval algorithm BM25 (43.23) and significantly outperforms the interpretable embeddings baseline CQG-MBQA (29.97). This shows that, although the relatedness scores are computed through general and symmetric semantic similarity, LDIR can still encode the unsymmetrical features of relationship between the query and the document in retrieval tasks.

We also notice that the traditional sparse representation method BM25 performs well in some tasks, such as achieving nDCG@10 results of 32.08 and 68.70 on NFC and SciFact, respectively. This indicates that embeddings based on sparse representation are still useful in some specific retrieval scenarios and have better interpretability. Nevertheless, our best averaged results are still slightly better than BM25 algorithm.

\begin{table*}[t!]
    \centering
    \scalebox{0.68}{
    \begin{tabular}{lrccrrrrrrrr}
        \toprule
        \multirow{2.5}*{\textbf{Model}} & \multirow{2.5}*{\textbf{Dim.}}  &\multirow{2.5}*{\textbf{Type}}&\multirow{2.5}*{\textbf{Perf.}}&\multicolumn{8}{c}{\textbf{Cognitive Load $\downarrow$}}\\
        \cmidrule(lr){5-12}
        
        &&& &\textbf{STS12} & \textbf{STS13} & \textbf{STS14} & \textbf{STS15} & \textbf{STS16} & \textbf{STS-B} & \textbf{SICK-R} & \textbf{Avg.}\\
        \midrule
        Bag-of-Words (with BERT vocabulary) & 28,996&Sparse&56.22 & 8& 4 & 6 & 5&  8& 7&6&6\\
        QAEmb-MBQA~\cite{benara2024crafting} &10,654 &0/1 Emb.&65.20 & 1,626& 1,571 & 1,625 & 1,443& 1,577 &1,408 &1,018&1,467\\
        CQG-MBQA~\cite{sun2024general} &9,614 &0/1 Emb.&77.60 &481 & 439 & 458 & 426& 478 & 446&413&449\\
        CQG-MBQA~\cite{sun2024general} &$^\dag$1,000 &0/1 Emb.&75.24 &48 &44 &43 &41 &44 &42 &37 &43\\
        \midrule
        LDIR (with AngIE, binarization) &500 &0/1 Emb. &75.87 &15 &13 &14 &12 &14 &13 &14 &14 \\
        LDIR (with AngIE, binarization) &200 &0/1 Emb. &71.34 &6 &5 &5 &4 &5 &5 &5 &5 \\
        \addlinespace[1pt]
        \cdashline{1-12}
        \addlinespace[2pt]
        LDIR (with AngIE, no binarization) &500 &Dense&82.82 &50 &52 &48 &47 &51 &43 &38 &47 \\
        LDIR (with AngIE, no binarization) &200 &Dense &81.99&15 &16 &15 &14 &15 &13 &12 &14 \\
        \bottomrule
    \end{tabular}}
\caption{Evaluation on cognitive load using STS tasks. $\dag$: The corresponding performance and cognitive load with 1000 dimensional embeddings are not provided and we estimate them from the curves in \citet{sun2024general}.}
\label{main_results_interpretability}
\end{table*}

\noindent\textbf{Clustering.} The results are shown in Table~\ref{main_results_clustering}. Different from STS and retrieval tasks, LDIR with SBERT$_\text{new}$ encoder gives the best results (39.36), outperforming the CQG-MBQA baseline (35.48) by 10.9\%. It also achieve comparable results to the embeddings by the backbone model SBERT$_\text{new}$ (38.87) and ModernBERT (41.55), validating the superior representation compactness of LDIR in unsupervised clustering. For ModernBERT and AngIE, the performance of LDIR with these encoders decrease, which may due to the transformation of relative representations in the low-dimensional space leads to a weakening of clustering characteristics for the high-dimensional representation space of these two embeddings.

\subsection{Evaluation on Cognitive Load}
We follow \citet{sun2024general} by using the ``cognitive load'' for measuring the interpretability. Formally, cognitive load is defined as the inner product of two binary embedding vectors $\mu$ and $\upsilon$:
\begin{equation}
\text{cognitive\ load} = \left\langle {\mu ,\upsilon } \right\rangle  = \sum\limits_{i = 1}^m {{\mu _i}{\upsilon _i}},
\label{cognitiveload}
\end{equation}
where a smaller cognitive load indicates stronger model interpretability, which means we only need to focus on fewer but more representative anchors to understand the text representation. This metric is originally used for binary embeddings and we transfer LDIR to binary embeddings by setting top-$k$ ($k=25$ according to the average) high values as 1 and the other values in different dimensions as 0.

The results are shown in Table~\ref{main_results_interpretability}. Firstly, bag-of-words gives a very cognitive load with 6, showing the highest interpretability. CQG-MBQA exhibits a cognitive load of 449 with 9000 dimensions, which reduces to 43 with 1000 dimensions, with a sacrifice of overall performance. Our LDIR with dimensions of 500 and 200, giving cognitive load values of 14 and 5 after binarization, respectively, showing a favorable performance (71.34$\sim$75.87) while maintaining a low cognitive load.

For LDIR without binarization, we find that it gives 47 and 14 cognitive load directly through Eq.~\ref{cognitiveload}, although this metric is not applicable to non-binary embeddings. Nevertheless, how to define the ``interpretability'' for text embeddings and design more suitable metrics that reflects interpretability for dense text embeddings still remains a valuable and open question.

\begin{figure}[t!]
    \centering
    \includegraphics[scale=0.33]{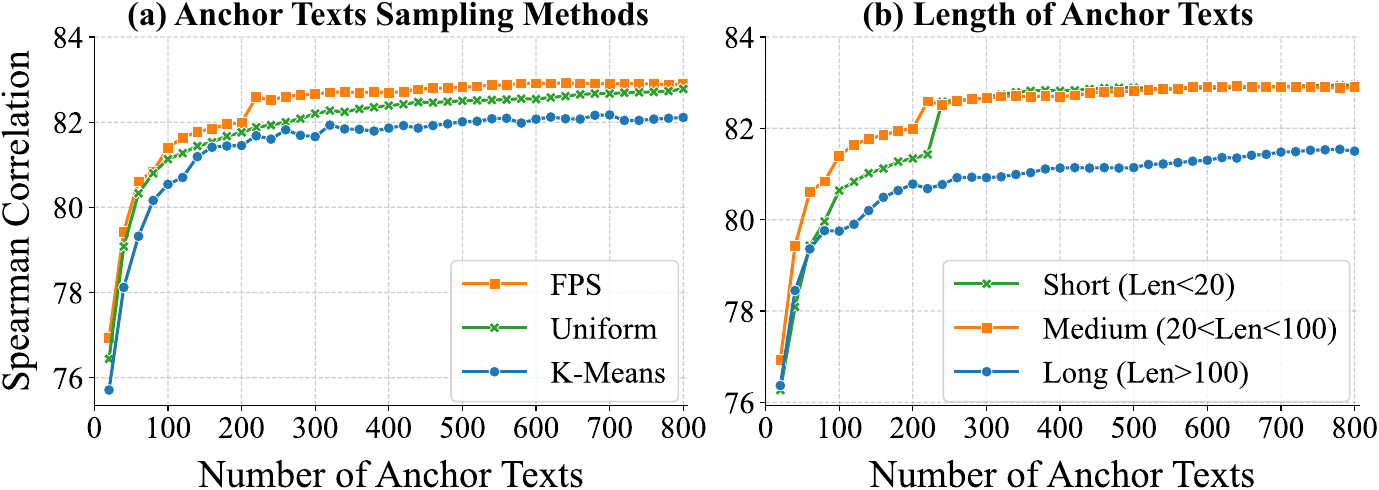}
    \caption{Comparison of different anchor texts sampling methods (a) and different settings of anchor texts length (b). The horizontal axis represents the different number (20$\sim$800) of different anchor texts. }
    \label{figure:merge}
\end{figure}

\section{Analyses and Discussions}

In our main experiments, we remove the long texts in corpus and use FPS to extract the anchor texts. We analysis the impact of sampling methods (Section \ref{compare_anchor_selection}) and the length of anchor texts (Section \ref{lengthofanchortexts}) below. Then we discuss how to obtain a more interpretable dense embeddings with the calculation of fine-grained relatedness (Section \ref{sec:relatedness}), compare different metrics on relatedness (Section \ref{sec:metrics}). Finally, we show a case study (Section \ref{casestudy}) and discuss the usage of relative representation based embeddings (Section \ref{sec:usage}).

\subsection{Anchor Texts Sampling Methods} 
\label{compare_anchor_selection}

We compare three anchor texts sampling methods: FPS, uniform sampling, and K-Means . The results are shown in Figure~\ref{figure:merge}(a). The trends of the three methods across different numbers of anchors were similar in general. FPS generally performed the best, with its performance converging with uniform sampling at 800 dimensions, while K-Means performed relatively poorly. The advantage of FPS in the low-dimensional range stems from its maximum diversity sampling strategy. The convergence of uniform sampling with FPS at 800 dimensions may be related to the properties of high-dimensional spaces when the dimensionality is sufficiently high, random uniform sampling can approximately cover the boundary regions of the semantic space. Notably, the K-Means method, constrained by the high similarity of cluster centers, resulted in insufficient semantic diversity in the anchor set compared to the other two methods, leading to suboptimal performance.

\begin{figure}[t!]
    \centering
    \includegraphics[scale=0.89]{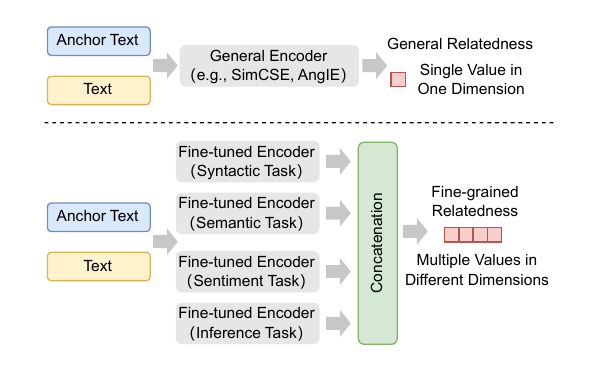}
    \caption{Comparison between the calculation of general (top) and fine-grained (bottom) relatedness scores. }
    \label{figure:finegrained}
\end{figure}

\begin{table}[t!]
    \centering
    \scalebox{0.63}{
    \begin{tabular}{lcccc}
        \toprule
       \textbf{Model}& \textbf{STS12} & \textbf{STS13} & \textbf{STS14} & \textbf{Avg.}\\
       \midrule
        Bag-of-Words (with BERT vocabulary) &44.75 &52.06 &54.78 &50.53\\
        QAEmb-MBQA~\cite{benara2024crafting} &59.40 &63.19 &57.68 &60.09\\
        CQG-MBQA~\cite{sun2024general}&69.21 &80.19 &73.91 &74.44\\
        
        \rowcolor{gray!20}
        LDIR (with fine-grained relatedness) &57.88 &67.33 &58.96 &61.39\\
        
        \bottomrule
    \end{tabular}}
\caption{The results of LDIR using the concatenated 800 dimensional embeddings with fine-grained relatedness.}
\label{finegrained_results}
\end{table}

\subsection{The Impact of Length of Anchor Texts} 
\label{lengthofanchortexts}
We investigate the impact of different length of anchor texts, including short (less than 20 tokens), medium (20$\sim$100 tokens), and long (longer than 100 tokens) texts. The results are shown in Figure~\ref{figure:merge}(b). We observe that for different settings of length, the performance improved rapidly using less than 200 anchor texts, after which the improvement is largely limited. The performance gap between short texts and medium texts narrowed and almost converged with 240 or more anchor texts, with both significantly outperforming long texts. We hypothesize that this may be attributable to the more intricate semantics and the potential inclusion of greater noise within longer texts in the corpus.

\subsection{Fine-grained Relatedness Calculation}
\label{sec:relatedness}

In Eq.~\ref{relcal}, the relatedness score is rather general, which reflects the overall semantic correlation between anchor text $a_j$ and text $t$. We further propose calculate more fine-grained relatedness score by applying different encoders tuned on different tasks, as shown in Figure~\ref{figure:finegrained} (bottom).

We use tasks in the general language understanding evaluation (GLUE; \citealp{wang-etal-2018-glue}) benchmark, which tests the text understanding ability from different perspectives. In particular, we select CoLA, STS-B, SST-2, and QNLI subtasks such that the fine-tuned models can encode syntactic, semantic, sentiment, and natural language inference features in fine-grained levels. Different models are fine-tuned on the training data and serve as encoders for calculating relatedness. The relatedness scores via 200 anchor texts are concatenated as a final 800-dimensional dense embeddings.

The results on STS12, STS13, and STS14 are shown in Table~\ref{finegrained_results}. We find that LDIR with fine-grained relatedness still outperforms the QAEmb-MBQA baseline (61.39 vs. 60.09). However, its overall performance decreases in STS tasks, underperforming the CQG-MBQA baseline (61.39 vs. 74.44). This show that, although the fine-grained relatedness scores have improved interpretability, their overall performance will not be enhanced in general semantic similarity tasks, which also reflects the trade-off between performance and interpretability, as discussed in Table~\ref{table:overall_comparison}.

\begin{table}[t!]
    \centering
    \scalebox{0.7}{
    \begin{tabular}{lcc}
        \toprule
       \textbf{
Metrics/Distance}& \textbf{Type} & \textbf{STS Results in Avg.} \\
       \midrule
       Edit Distance&surface-based&	17.99\\
        Jaccard	&surface-based	&22.34\\
        \midrule
Sokalsneath	&vector-based \& binary	&79.04\\
Jaccard	&vector-based \& binary	&79.49\\
Hamming	&vector-based \& binary	&80.00\\
Dice	&vector-based \& binary	&80.07\\
\midrule
    Chebyshev&	vector-based \& dense&	79.33\\
    Euclidean&	vector-based \& dense&	80.76\\
        Manhattan	&vector-based \& dense	&81.19\\
        \rowcolor{gray!20}
       Cosine (this work) &vector-based \& dense &82.82 \\
        
        \bottomrule
    \end{tabular}}
\caption{The results of LDIR using different surface-based and vector-based metrics.}
\label{metrics}
\end{table}

\subsection{Surface-based and Vector-based Metrics}
\label{sec:metrics}
We further try surface-based metrics as well as other various vector-based metrics for calculating the relatedness, and the results for 500 dimensional embeddings are shown in Table~\ref{metrics}. We find that surface-based distances cannot effectively represent semantic relationships, where the results are very poor. On the other hand, other vector-based distance (even when applied to LDIR after binarization) consistently show close results, despite differences in distances calculation against the cosine distances in downstream tasks.

\subsection{Case Study}
\label{casestudy}
Table~\ref{table:casestudy} shows a case of four-dimensional LDIR embeddings of two title texts from SCIDocs. Text A has the highest relevance to anchor text 1, while text B is more related to anchor texts 3 and 4, resulting in the difference in their LDIR embeddings.

\subsection{The Usage of Relative Representation}
\label{sec:usage}
Our approach is based on transforming relative representations using an existing backbone model, where we consider its potential applications here briefly.  For instance, some studies have shown that it is possible to recover original training data or sensitive information from embeddings themselves~\cite{morris-etal-2023-text,li-etal-2023-sentence,huang-etal-2024-transferable}. Therefore, directly exposing embedding interfaces may lead to data privacy or leakage issues. Consider a scenario where an embedding provider (such as text-embedding-Ada series by OpenAI) does not want directly provide users with high-dimensional raw embeddings outputs. Instead, they could offer low-dimensional relative representation vectors based on anchor text (which can be provided by the provider or defined by the user). This approach 1) can help avoid some data privacy or leakage concerns; 2) performs well in certain downstream applications; 3) allows for obtaining more text representations with the same storage and computational cost with low-dimensionality; 4) provides interpretability and reliability with the flexibility of anchor text.

\begin{table}[t!]
	\centering
    \scalebox{0.47}{
    \begin{tabular}{lcccc}
	    \toprule 
        
        \multirow{2}{*}{\textbf{Anchor Text \#1}:} & \multicolumn{4}{l}{A novel, high-performing architecture for end-to-end named entity recognition and}\\
    &\multicolumn{4}{l}{relation extraction that is fast to train.}\\
    \midrule
    \multirow{2}{*}{\textbf{Anchor Text \#2}:} & \multicolumn{4}{l}{Predicting affective states expressed through an emote-aloud procedure from auto-}\\
    &\multicolumn{4}{l}{tutor's mixed-initiative dialogue.}\\
    \midrule
    \multirow{2}{*}{\textbf{Anchor Text \#3}:} & \multicolumn{4}{l}{Planar High-Gain Dielectric-Loaded Antipodal Linearly Tapered Slot Antenna for }\\
        &\multicolumn{4}{l}{$E$- and $W$-Band Gigabyte Point-to-Point Wireless Services}\\
        \midrule
    \multirow{2}{*}{\textbf{Anchor Text \#4}:} & \multicolumn{4}{l}{Design of a Monopulse Antenna Using a Dual V-Type Linearly Tapered Slot Ant-}\\
    &\multicolumn{4}{l}{enna (DVLTSA).}\\
\end{tabular}}

\scalebox{0.477}{
\begin{tabular}{llcccc}
\midrule 
    \multicolumn{2}{l}{\textbf{Texts ($\downarrow$) and LDIR Embeddings ($\rightarrow$)} }&\textbf{Dim \#1}&\textbf{Dim \#2}&\textbf{Dim \#3}&\textbf{Dim \#4}\\
    \midrule
    \multirow{2}{*}{\textbf{Text A}:} &Morphological Embeddings for Named Entity &\multirow{2}{*}{0.7554} &\multirow{2}{*}{0.5312} &\multirow{2}{*}{0.3801}  &\multirow{2}{*}{0.3607}\\
    &Recognition in Morphologically Rich Languages. &&&&\\
    \midrule

    \multirow{2}{*}{\textbf{Text B}:} &Design Approach to a Novel Dual-Mode Wide- &\multirow{2}{*}{0.3941} &\multirow{2}{*}{0.3373} &\multirow{2}{*}{0.6874}  &\multirow{2}{*}{0.6561}\\
    & band Circular Sector Patch Antenna. &&&&\\
    	\bottomrule
	\end{tabular}}
	\caption{LDIR embeddings of text A and text B. We only show values on four dimensions for brevity.}
	\label{table:casestudy}
\end{table}

\section{Conclusion}
We propose LDIR, a low-dimensional dense and interpretable text embeddings with relative representations. The numerical values in LDIR represent the correlation between the text and the anchor texts automatically obtained through farthest point sampling, which improves the expressiveness at a low cost compared with QA-based 0/1 embeddings. Across multiple tasks and datasets, LDIR demonstrates better semantic expressiveness compared to multiple interpretable embedding baselines. We also discuss the need for more suitable interpretability metrics and how to further enhance the interpretability of LDIR, which can serve as our future work for interpretable text embeddings.

\section*{Limitations}
Interpretable text embeddings is a relatively new research topic and the baselines of interpretable text embeddings we used is limited. Also, as we mentioned, there is still room for improvement in the interpretability of our embeddings. According to our approach, the anchor texts are sampled from the entire corpus, and we can further optimize these anchor texts for different downstream tasks for better task-related expressiveness. Additionally, the human evaluation and application scenarios of interpretable text embeddings can be further explored, which is also crucial for explainable AI.


\bibliography{custom}

\appendix
\section{Checkpoints of Baseline Models}
\label{appendix:resources}
In Table~\ref{resources}, we list the checkpoints of baseline models we used in experiments.

\begin{table*}[h!]
\centering
    \resizebox{0.9\linewidth}{!}{
    \begin{tabular}{ll}
    \toprule 
    \textbf{Model}&\textbf{Checkpoint}\\
         \midrule
         BERT&\url{https://huggingface.co/google-bert/bert-base-uncased}\\
         GloVe&\url{https://huggingface.co/sentence-transformers/average_word_embeddings_glove.6B.300d}\\
         SimCSE$_\text{unsup}$&\url{https://huggingface.co/princeton-nlp/sup-simcse-bert-base-uncased}\\
         SimCSE$_\text{sup}$&\url{https://huggingface.co/princeton-nlp/unsup-simcse-bert-base-uncased}\\
         WhitenedCSE&\url{https://huggingface.co/SupstarZh/whitenedcse-bert-base}\\
         SBERT$_\text{ori}$&\url{https://huggingface.co/sentence-transformers/all-mpnet-base-v1}\\
         SBERT$_\text{new}$&\url{https://huggingface.co/sentence-transformers/all-mpnet-base-v2}\\
         ModernBERT$_\text{emb}$&\url{https://huggingface.co/lightonai/modernbert-embed-large} \\
         text-embedding-ada-002&\url{https://openai.com/index/new-and-improved-embedding-model/}\\
         AngIE&\url{https://huggingface.co/WhereIsAI/UAE-Large-V1}\\
         LLM2Vec&\url{https://huggingface.co/McGill-NLP/LLM2Vec-Meta-Llama-3-8B-Instruct-mntp-unsup-simcse}\\
         BM25&\url{https://github.com/xhluca/bm25s}\\
        \bottomrule
    \end{tabular}}
    \caption{Checkpoints of baseline models we used in experiments.}
    \label{resources}
\end{table*} 

\end{document}